\pdfoutput=1
\PassOptionsToPackage{table,dvipsnames}{xcolor}
\documentclass[11pt,letterpaper,logo,onecolumn]{main}
\usepackage[numbers,sort&compress,square]{natbib}

\usepackage{microtype}

\usepackage{enumitem}
\usepackage{needspace}

\usepackage{amsmath,amssymb,amsthm}
\usepackage{mathtools}
\usepackage{bm}

\usepackage{amsmath,amsfonts,bm}

\def\eqref#1{equation~\ref{#1}}

\def\1{\bm{1}}

\DeclareMathAlphabet{\mathsfit}{\encodingdefault}{\sfdefault}{m}{sl}
\SetMathAlphabet{\mathsfit}{bold}{\encodingdefault}{\sfdefault}{bx}{n}

\newcommand{\E}{\mathbb{E}}

\newcommand{\Var}{\mathrm{Var}}

\newcommand{\Cov}{\mathrm{Cov}}

\usepackage{algorithm}
\usepackage[noend]{algpseudocode}

\usepackage{graphicx}
\usepackage{float}
\usepackage{placeins}

\usepackage{booktabs}
\usepackage{multirow}
\usepackage{array}
\usepackage{tabularx}
\usepackage{etoolbox}
\usepackage{fontawesome}
\usepackage[most]{tcolorbox}
\definecolor{gblue9}{RGB}{23,78,166}
\definecolor{bandgray}{HTML}{F1F3F5}
\definecolor{oursrow}{HTML}{EAF5F2}
\definecolor{oursteal}{HTML}{0F6B61}

\newcommand{\sd}[1]{{\scriptsize\textcolor{black!55}{$\pm$#1}}}
\newcommand{\oursrow}{\rowcolor{oursrow}}
\newcommand{\gain}[1]{{\footnotesize\textcolor{oursteal}{$+$#1}}}
\newcommand{\gainrow}{{\footnotesize\textcolor{oursteal}{\quad$\Delta$ vs.\ \rar}}}

\newtcolorbox[auto counter]{findingbox}{%
  enhanced, colback=oursrow, colframe=oursteal, boxrule=0pt, leftrule=2.2pt,
  arc=1mm, outer arc=1mm, left=8pt, right=8pt, top=4pt, bottom=4pt,
  before skip=7pt, after skip=5pt, fontupper=\bfseries}
\makeatletter
\newcommand{\finding}[1]{%
  \Needspace{5\baselineskip}%
  \begin{findingbox}
    {\color{oursteal}\faLightbulbO\enspace Finding~\thetcbcounter.}\enspace#1
  \end{findingbox}%
  \@afterindentfalse\@afterheading}
\makeatother

\AtBeginEnvironment{table}{\setlength{\belowcaptionskip}{5pt}}
\AtBeginEnvironment{figure}{%
  \setlength{\abovecaptionskip}{6pt}%
  \setlength{\belowcaptionskip}{0pt}}

\fancypagestyle{titlefolio}[firststyle]{%
  \fancyfoot[L]{\footerfont
                The full author list appears at the end of the paper.}%
  \fancyfoot[R]{\footerfont\thepage}}

\hypersetup{colorlinks=true, linkcolor=black, citecolor=blue, urlcolor=blue}

\usepackage{xspace}
\newcommand{\method}{\textsc{RAR}\textsuperscript{2}\xspace}
\newcommand{\rar}{RaR\xspace}
\newcommand{\arl}{\textsc{ARL}\xspace}
\newcommand{\powthree}{\textsc{POW3R}\xspace}
\newcommand{\onemillion}{\$OneMillion-Bench\xspace}
\newcommand{\Corr}{\operatorname{Corr}}
\newcommand{\Prob}{\mathbb{P}}
\newcommand{\ind}{\mathbb{I}}

\newcommand{\brandiconraise}[1]{\raisebox{-0.18\height}{#1}}
\makeatletter
\newrobustcmd{\brandicon}[1]{\@nameuse{brandicon@#1}}
\@namedef{brandicon@openai}{%
  \brandiconraise{\includegraphics[height=0.92em]{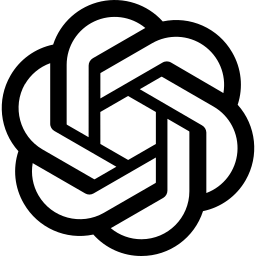}}}
\@namedef{brandicon@claude-ai}{%
  \brandiconraise{\includegraphics[height=0.92em]{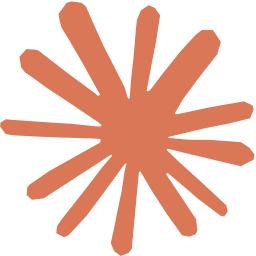}}}
\@namedef{brandicon@hunyuan-color}{%
  \brandiconraise{\includegraphics[height=0.92em]{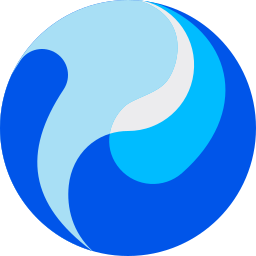}}}
\@namedef{brandicon@deepseek-color}{%
  \brandiconraise{\includegraphics[height=0.92em]{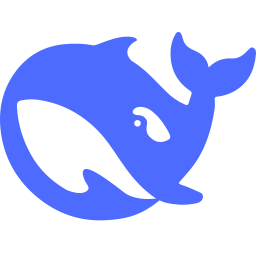}}}
\@namedef{brandicon@qwen-color}{%
  \brandiconraise{\includegraphics[height=0.92em]{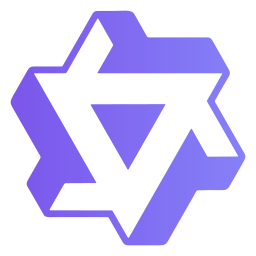}}}
\makeatother
\newrobustcmd{\modelname}[2]{\mbox{\brandicon{#1}\hspace{0.32em}#2}}
\newcommand{\groupband}[2]{%
  \rowcolor{bandgray}\multicolumn{#1}{@{}l@{}}{\textbf{#2}}}
\newcommand{\policyband}[2]{%
  \rowcolor{bandgray}\multicolumn{#1}{@{}l@{}}{%
    \brandicon{qwen-color}\hspace{0.42em}\textbf{#2}%
    \quad\textcolor{black!55}{\textit{policy model}}}}

\newcommand{\authorname}[1]{\textbf{#1}}
\newcommand{\myheaderbreak}{\\}
\title{\texorpdfstring{\textsc{RAR}\textsuperscript{2}}{RAR2}: Reliability-Aware
  Rubrics as Rewards\myheaderbreak for Reinforcement Learning}

\author{
  \normalsize
  Hunyuan Team, Tencent
  \qquad
  Fudan University\\
  \textbf{University of Hong Kong}
  \qquad
  \textbf{University of Toronto}
}

\newcommand{\fullauthorlist}{%
  \section*{Full Author List}
  \authorname{Zhangyue Yin}\textsuperscript{1},
  \authorname{Yufei Huang}\textsuperscript{1},
  \authorname{Xiong Kuang}\textsuperscript{1},
  \authorname{Zongwei Li}\textsuperscript{1},
  \authorname{Hao Ma}\textsuperscript{1},
  \authorname{Jirui Yang}\textsuperscript{1,2},
  \authorname{Boyang Liu}\textsuperscript{1,2},
  \authorname{Qiushi Sun}\textsuperscript{3},
  \authorname{Helena Shunhua Zhao}\textsuperscript{4},
  \authorname{Ming Zhang}\textsuperscript{1,2},
  \authorname{Shihan Dou}\textsuperscript{1,2},
  \authorname{Tao Gui}\textsuperscript{2},
  \authorname{Xuanjing Huang}\textsuperscript{2},
  \authorname{Jianfeng Yan}\textsuperscript{1},
  \authorname{Fengzong Lian}\textsuperscript{1}

  \vspace{0.35em}
  \par\noindent
  \textsuperscript{1}Hunyuan Team, Tencent\quad
  \textsuperscript{2}Fudan University\quad
  \textsuperscript{3}University of Hong Kong\quad
  \textsuperscript{4}University of Toronto}

\begin{abstract}
LLMs are increasingly deployed across open-ended tasks, where high-quality
outputs must satisfy multiple, task-specific criteria. Rubric-based evaluation
makes these criteria explicit and is used to assess complex
outputs in specialized, high-stakes domains.
The Rubrics as Rewards (\rar) framework extends this approach to reinforcement
learning by aggregating rubric-level verdicts into structured reward signals.
We characterize rubric signals along two dimensions: discrimination across
policy-generated responses and stability across repeated evaluations. The two
dimensions divide rubric signals into four quadrants, with stable and
discriminative signals providing the most reliable guidance for policy
learning. However, standard \rar implicitly assumes noise-free rubric verdicts
and therefore cannot distinguish genuine differences between responses from
those induced by unstable judgments.
To address this limitation, we propose \emph{Reliability-Aware Rubrics as
Rewards} (\method). \method measures stability through within-response
variance and discrimination through between-response variance. A
Beta--Binomial random-effects model then assesses whether each rubric's
apparent discrimination is supported by stable judgments, and a reliability
gate filters out rubric signals that lack such support.
Experiments show that \method achieves the best performance across models of
different sizes and diverse tasks.
Additional analyses demonstrate that \method delivers consistent performance
gains across judge reasoning-effort settings and improves agreement with
expert assessments across judge models.
\end{abstract}

\begin{document}
\raggedbottom
\clubpenalty=10000
\widowpenalty=10000
\makeatletter\@clubpenalty=10000\makeatother
\setlength{\parskip}{0.3\baselineskip}
\maketitle
\thispagestyle{titlefolio}
\renewcommand{\myheaderbreak}{ }
\makeatletter\global\@topnum\z@\makeatother

\section{Introduction}
\label{sec:introduction}

As large language models (LLMs) are increasingly deployed in open-ended
applications, evaluating response quality across multiple dimensions has
become essential
\citep{DBLP:journals/corr/abs-2211-09110,DBLP:conf/emnlp/LaskarABRKKJB0P24}.
Reliable evaluation is particularly important in medicine, law, and finance,
where domain-specific standards govern response quality and errors may have
serious consequences
\citep{DBLP:journals/corr/abs-2505-08775,DBLP:conf/acl/AkyurekGZGJGRMR26}.
Reducing such multidimensional quality to a
single holistic verdict obscures which requirements are met and leaves the
judgment vulnerable to known biases and inconsistencies
\citep{DBLP:conf/nips/ZhengC00WZL0LXZ23,
DBLP:journals/corr/abs-2310-10076,DBLP:journals/corr/abs-2606-13685}.
Rubric-based evaluation instead decomposes quality into explicit criteria,
yielding criterion-level judgments that are more transparent and actionable
\citep{DBLP:conf/acl/HashemiERDK24,DBLP:journals/corr/abs-2602-05125}.
Prompt-specific rubrics have been used to assess medical and professional reasoning
\citep{DBLP:journals/corr/abs-2505-08775,DBLP:conf/acl/AkyurekGZGJGRMR26}. The Rubrics as Rewards
framework (\rar; \citealp{DBLP:journals/corr/abs-2507-17746}) turns their criterion-level
verdicts into structured, interpretable reward signals for on-policy
reinforcement learning.

However, the effectiveness of rubric scores as reinforcement-learning rewards
can be characterized along two complementary dimensions: judgment stability
and response discrimination. Judgment stability quantifies the consistency of
evaluations of the same response, whereas response discrimination
captures the extent to which a rubric distinguishes among responses sampled
from the current policy. Figure~\ref{fig:variance_counterexample}
highlights the importance of considering these dimensions jointly. Although
response discrimination is essential for a rubric to yield an informative
learning signal, it is insufficient unless the underlying judgments are also
stable.
For unstable rubrics, apparent discrimination may reflect judge noise rather
than genuine response-dependent variation, thereby distorting credit assignment
and impairing policy optimization. This two-dimensional characterization
motivates our central research question:

\begin{figure}[t]
    \centering
    \includegraphics[width=0.95\linewidth]{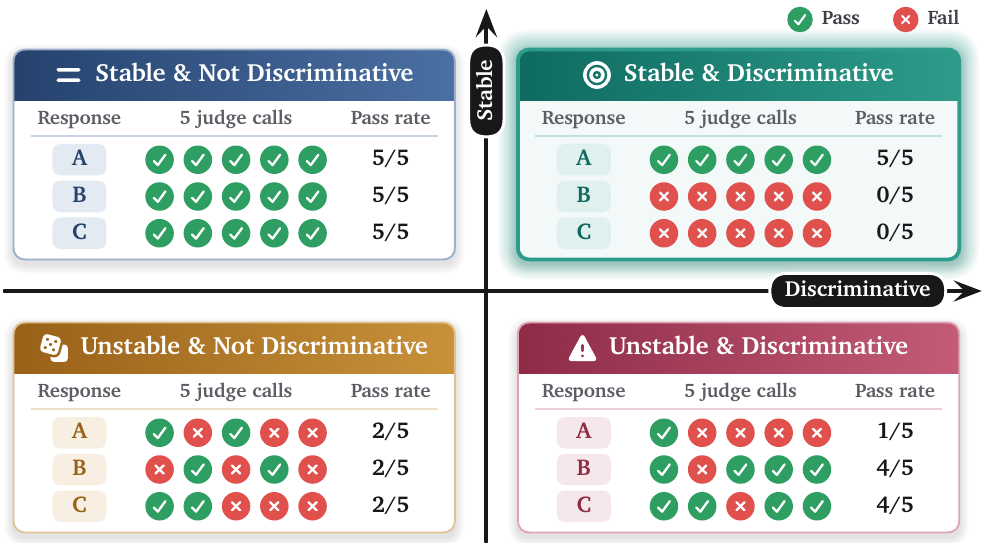}
    \caption{\textbf{A useful rubric reward needs both stability and
    discrimination.} Each row fixes one response and repeats the same rubric
    judgment five times. Across-response differences provide discrimination;
    within-response agreement establishes stability. Only the upper-right
    quadrant supports reliable credit assignment across responses. \method
    targets this stable, discriminative signal while filtering apparent contrast caused by
    inconsistent judgments.}
    \label{fig:variance_counterexample}
\end{figure}

\begin{quote}
\emph{How can we provide stable and discriminative rubric-based reward signals
for reinforcement learning?}
\end{quote}

To address this challenge, we introduce \emph{Reliability-Aware Rubrics as
Rewards} (\method), a novel method for selecting rubric signals that are both
stable and discriminative. Leveraging multiple evaluations under a fixed judge
protocol, \method assesses judgment stability from within-response variance
and response discrimination from between-response variance. Because direct
variance estimates are imprecise for small response groups, a Beta--Binomial
random-effects model estimates the posterior reliability of each rubric's
response contrast. \method retains rubric signals whose discrimination is supported by
stable judgments and filters unreliable signals from the training reward.
The gate preserves the original weights of retained rubrics and defaults to
retention when statistical evidence is insufficient. Moreover, \method
exploits the conditional independence of repeated judgments to query the judge
in parallel.

\method delivers consistent gains across multiple medical and
professional reasoning benchmarks, substantially outperforming a suite of
strong baselines. Further analyses demonstrate that these improvements persist
across policy scales and judge reasoning-effort settings, while the resulting
reliability-aware signals exhibit stronger agreement with expert assessments
across judge models.

Our contributions are:
\begin{enumerate}[leftmargin=1.45em,itemsep=4pt,topsep=3pt]
    \item We propose \emph{Reliability-Aware Rubrics as Rewards} (\method), a
    novel method that addresses unreliable rubric-based rewards by retaining
    learning signals that are both stable and discriminative.
    \item We characterize stability through within-response variance and
    discrimination through between-response variance, using a Beta--Binomial
    random-effects model to estimate their contributions.
    \item We demonstrate that \method consistently outperforms strong baselines
    across multiple benchmarks and improves agreement between rubric-derived
    learning signals and expert assessments across judge models.
\end{enumerate}

\section{Preliminaries}
\label{sec:preliminaries}

To establish the formal basis for \method, we first formalize rubric-based
reward construction for reinforcement learning and introduce the associated
notation. We then decompose rubric-level variation into stable differences
across responses and within-response judgment noise, providing the basis for
our definition of rubric reliability.

\subsection{Rubrics as Rewards}

Fix a prompt $x$ and let
$\mathcal R=\{r_1,\ldots,r_M\}$ denote its set of $M$ rubrics. Each rubric
$r\in\mathcal R$ has an importance weight $w_r$. For a response $y_g$, let
$c_r(x,y_g)$ denote its score under rubric $r$. Rubrics as Rewards
(\rar; \citealp{DBLP:journals/corr/abs-2507-17746}) aggregates these
rubric-level scores as
\begin{equation}
S_g^{\mathrm{RaR}}
:=\frac{1}{W}\sum_{r\in\mathcal R}w_r c_r(x,y_g),
\qquad
W:=\sum_{r\in\mathcal R}\max(w_r,0)>0.
\label{eq:rar_baseline_score}
\end{equation}
When all weights are nonnegative, $W=\sum_{r\in\mathcal R}w_r$. If negative
weights are present, they contribute to the weighted numerator but are excluded
from the normalization denominator. This normalization makes rewards comparable
across prompts with different rubric counts or weights.

\subsection{Multiple evaluations and variance decomposition}

To estimate judgment stability, \method evaluates $G$ responses
$y_1,\ldots,y_G$ under each rubric $K$ times using a fixed judge protocol. We
use $g$ for responses, $k$ for evaluations, and $r$ for rubrics. Let
$v_{gk,r}\in\{0,1\}$ denote the $k$-th verdict on rubric $r$ for response
$y_g$, and collect the verdicts of one rubric in
\begin{equation}
V_r:=[v_{gk,r}]_{g=1,\ldots,G;\,k=1,\ldots,K}
\in\{0,1\}^{G\times K},
\qquad
\bar v_{g,r}:=\frac{1}{K}\sum_{k=1}^{K}v_{gk,r}.
\label{eq:mean_verdict}
\end{equation}
When $r$ is clear from context, we write $v_{gk}$ and $\bar v_g$ for
$v_{gk,r}$ and $\bar v_{g,r}$. The rubric score in
Eq.~\ref{eq:rar_baseline_score} is this mean verdict,
$c_r(x,y_g)=\bar v_{g,r}$, so \rar is recovered from the gated score of
Section~\ref{sec:selecting_rubrics} when every rubric is retained.

For the fixed rubric, let $q_g\in[0,1]$ denote the latent judge pass
probability for response $y_g$. It is the probability that the fixed judge
protocol returns a positive verdict for that response. Conditional on $q_g$,
the repeated verdicts $v_{g1},\ldots,v_{gK}$ are independent Bernoulli draws.
Across responses, the law of total variance gives
\begin{equation}
\Var(v_{gk})=
\underbrace{\Var(q_g)}_{\text{stable response contrast}}+
\underbrace{\E[q_g(1-q_g)]}_{\text{within-response judge noise}}.
\label{eq:variance_decomp}
\end{equation}
With $K=1$, the observed variance contains only this sum, so a large
rollout-level standard deviation is not evidence of reproducible contrast.
Multiple judgments hold the response fixed and expose judge instability.
Comparing their means $\bar v_g$ then reveals stable differences across
responses. Estimating this variance decomposition directly is unstable for the
small response groups used during training. Appendix~\ref{app:derivations}
provides the derivation.

\section{Reliability-Aware Rubrics as Rewards}
\label{sec:method}

\begin{figure}[t]
\centering
\includegraphics[width=\linewidth]{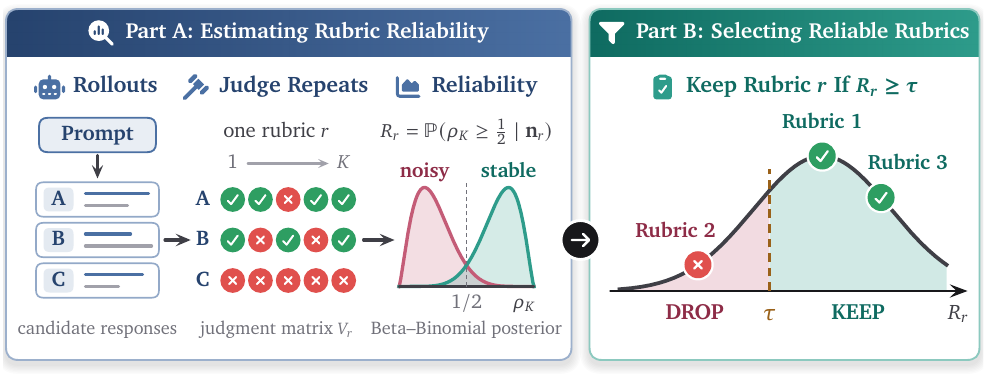}
\caption{\textbf{Overview of \method.}
\textbf{Part A: Estimating Rubric Reliability.} Multiple evaluations determine whether each rubric
produces stable response contrast or noisy judge decisions.
\textbf{Part B: Selecting Reliable Rubrics.} Rubrics are positioned by $R_r$,
the posterior probability that $\rho_K\ge\tfrac12$; rubrics with $R_r<\tau$
are dropped, while the rest are retained for reward construction.}
\label{fig:method_overview}
\end{figure}

\emph{Reliability-Aware Rubrics as Rewards} (\method) addresses this challenge
by estimating whether a rubric's observed contrast reflects stable differences
among responses rather than variation in the judge's verdicts. \method proceeds in two stages: (i) estimate each
rubric's reliability from multiple evaluations and (ii) select reliable rubrics
to construct the reliability-aware training reward.

\subsection{Estimating Rubric Reliability}
\label{sec:estimating_reliability}

Fix a rubric $r$ and use the entries $v_{gk}$ of its verdict matrix $V_r$.
Summarize the $K$ verdicts for response $y_g$ by its positive-verdict count
\begin{equation}
n_g:=\sum_{k=1}^{K}v_{gk},
\qquad
\mathbf n:=(n_1,\ldots,n_G).
\label{eq:rubric_counts}
\end{equation}
Thus one rubric is represented by a length-$G$ count vector rather than a
three-index verdict tensor.

For $G\ge2$ and $K\ge2$, \method fits each rubric independently with a
Beta--Binomial random-effects model
\citep{skellam1948variablebinomial,prentice1986extendedbetabinomial}:
\begin{align}
q_g\mid\mu,\phi
&\overset{\mathrm{iid}}{\sim}
\mathrm{Beta}\!\left(\mu\kappa,(1-\mu)\kappa\right),
\qquad \kappa=\phi^{-1}-1,\nonumber\\
n_g\mid\mu,\phi
&\overset{\mathrm{ind}}{\sim}
\mathrm{BetaBinomial}
\!\left(K,\mu\kappa,(1-\mu)\kappa\right).
\label{eq:betabin_model}
\end{align}
Here $\mu$ is the rubric's mean judge pass probability across responses, while
$\phi$ is the intraclass correlation between repeated
verdicts for the same response. Under this parameterization, the concentration
$\kappa$ is fully specified by $\phi$ rather than estimated independently.
Every rubric has its own
$(\mu,\phi)$ pair, and we keep omitting the index $r$ because each fit
concerns a single rubric.

For the fixed ensemble size $K$, define rubric reliability as the fraction
of variance in $\bar v_g$ attributable to stable differences between
responses:
\begin{equation}
\rho_K
:=\frac{\Var(q_g)}{\Var(\bar v_g)}
=\frac{K\phi}{1+(K-1)\phi}.
\label{eq:rubric_reliability}
\end{equation}
This is the Spearman--Brown reliability of an average of $K$ exchangeable
measurements when the single-measure intraclass correlation is $\phi$
\citep{shrout1979icc,spearman1910correlation}.

\paragraph{Variance decomposition and interpretation.}
The model separates stable response variation
$\sigma_B^2:=\Var(q_g)=\mu(1-\mu)\phi$ from within-response judge noise
$\sigma_W^2:=\E[q_g(1-q_g)]=\mu(1-\mu)(1-\phi)$.
Averaging $K$ conditionally independent verdicts leaves $\sigma_B^2$ unchanged
and divides $\sigma_W^2$ by $K$, giving
$\Var(\bar v_g)
=\sigma_B^2+\frac{\sigma_W^2}{K}
=\mu(1-\mu)
\left(\phi+\frac{1-\phi}{K}\right)$. Hence, under
Eq.~\ref{eq:betabin_model}, $\rho_K$ in Eq.~\ref{eq:rubric_reliability} is the
fraction of mean-verdict variance explained by stable response differences;
Appendix~\ref{app:derivations} gives the full derivation.

\paragraph{Identifiability from multiple evaluations.}

When $K=1$, the count $n_g$ is binary and its marginal likelihood is
\begin{equation}
p(n_g\mid\mu,\phi)
=\mu^{n_g}(1-\mu)^{1-n_g}.
\label{eq:single_judgment_likelihood}
\end{equation}
Equation~\ref{eq:single_judgment_likelihood} contains no information about
$\phi$. Rubric reliability therefore cannot be estimated from one judgment
per response. When $K\ge2$, the count variance is
\begin{equation}
\Var(n_g\mid\mu,\phi)
=K\mu(1-\mu)\left[1+(K-1)\phi\right].
\label{eq:count_variance}
\end{equation}
Larger values of $\phi$ induce greater overdispersion in the response-level
counts, reflecting stronger agreement among judgments of the same response.
Accordingly, $\phi$ is identifiable from the count distribution when $K\ge2$.

\paragraph{Effect of multiple evaluations.}

For fixed $0<\phi<1$, $\rho_K$ increases monotonically with $K$ and converges
to one as $K\to\infty$. More judgments therefore reduce the contribution of
judge noise to the mean verdict but increase evaluation cost. In practice,
$K$ should balance this cost against the reliability required for a stable
learning signal.

\begin{algorithm}[t]
\caption{Reliability-Aware Rubrics as Rewards}
\label{alg:rar}
\small
\begin{algorithmic}[1]
\Require prompt $x$, policy $\pi_\theta$, rubric set
$\mathcal R=\{r_1,\ldots,r_M\}$ with weights $w_r$, group size $G$,
judgments per response $K$, threshold $\tau$
\State sample responses $y_1,\ldots,y_G\sim\pi_\theta(\cdot\mid x)$
\State collect binary verdicts $v_{gk,r}$ for every response, judgment, and rubric
\ForAll{$r\in\mathcal R$}
    \State $n_{g,r}\gets\sum_{k=1}^{K}v_{gk,r}$ for $g=1,\ldots,G$
    \State $\mathbf n_r\gets(n_{1,r},\ldots,n_{G,r})$
    \State $R_r\gets\Prob(\rho_K\ge\tfrac12\mid\mathbf n_r)$ under the
    Beta--Binomial model
    \If{evidence is insufficient or domain policy protects $r$}
        \State $m_r\gets1$
    \Else
        \State $m_r\gets\ind[R_r\ge\tau]$
    \EndIf
\EndFor
\State $W_m\gets\sum_{r\in\mathcal R}m_r\max(w_r,0)$
\If{$W_m=0$}
    \State $m_r\gets1$ for every $r\in\mathcal R$
    \State $W_m\gets\sum_{r\in\mathcal R}\max(w_r,0)$
\EndIf
\For{$g=1,\ldots,G$}
    \State $S_g\gets\frac{1}{K W_m}
    \sum_{r\in\mathcal R}w_r m_r n_{g,r}$
\EndFor
\State $\bar S\gets\frac{1}{G}\sum_{g=1}^{G}S_g$
\State update $\pi_\theta$ from the relative credits $\{S_g-\bar S\}_{g=1}^{G}$
\Statex \textbf{Evaluation:} skip gating by setting $m_r=1$ for all
$r\in\mathcal R$.
\end{algorithmic}
\end{algorithm}

We place independent priors
$\mu\sim\mathrm{Beta}(1,1)$ and $\phi\sim\mathrm{Beta}(1,4)$.
Combining these priors with the Beta--Binomial likelihood yields the posterior
\begin{equation}
p(\mu,\phi\mid\mathbf n)
\propto
p(\mu)p(\phi)
\prod_{g=1}^{G}p(n_g\mid\mu,\phi).
\label{eq:posterior}
\end{equation}
The prior on $\phi$ places greater mass on small values,
so high reliability requires clear evidence from multiple judgments of stable
variation across responses. For each rubric $r$, the resulting posterior is
summarized by the gate statistic
\begin{equation}
R_r
:=\Prob(\rho_K\ge\tfrac12\mid\mathbf n_r).
\label{eq:posterior_summaries}
\end{equation}
$R_r$ is the posterior probability that stable between-response differences
account for at least half of the variance in mean verdicts under the fixed
judge protocol.

We assume the same number of valid judgments for every response.
Algorithm~\ref{alg:rar_full} in Appendix~\ref{app:implementation} describes
how these judgments are collected through parallel judge calls.

\subsection{Selecting Reliable Rubrics}
\label{sec:selecting_rubrics}

Let $\tau\in(0,1)$ be the required posterior confidence and let
$m_r\in\{0,1\}$ indicate whether rubric $r$ is retained. \method uses
\begin{equation}
m_r=
\begin{cases}
1,&\text{if a fail-open safeguard applies},\\
\ind[R_r\ge\tau],&\text{otherwise}.
\end{cases}
\label{eq:gate}
\end{equation}
Fail-open safeguards retain a rubric when evidence is insufficient or when the
rubric is protected for normative reasons.
The threshold $\tau$ controls the tradeoff between retaining noisy rubrics and
discarding reliable rubrics.
Because $\rho_K=\sigma_B^2/(\sigma_B^2+\sigma_W^2/K)$, the event
$\rho_K\ge\tfrac12$ is equivalent to $\sigma_B^2\ge\sigma_W^2/K$. The gate
therefore asks whether a rubric's between-response contrast is at least as
large as the judge noise that remains after averaging $K$ verdicts, combining
stability and discrimination in a single criterion, as detailed in
Appendix~\ref{app:gate_geometry}.

\paragraph{Reliability-aware reward construction.}

For illustration, consider a group-relative update such as GRPO
\citep{DBLP:journals/corr/abs-2402-03300}, which assigns credit by comparing
rewards within the $G$-response prompt group.
For each rubric $r$, define its group-mean verdict as
\begin{equation}
\bar v_{\cdot,r}
:=\frac{1}{G}\sum_{j=1}^{G}\bar v_{j,r}.
\label{eq:group_mean_verdict}
\end{equation}
Let $W_m:=\sum_{r\in\mathcal R}m_r\max(w_r,0)$ denote the retained positive
weight mass. If $W_m=0$, \method fails open by setting $m_r=1$ for every
$r\in\mathcal R$.
The reliability-gated score and the centered contrast used for relative credit
are
\begin{equation}
S_g=
\frac{1}{W_m}\sum_{r\in\mathcal R}w_r m_r\bar v_{g,r},
\qquad
S_g-\bar S
=\frac{1}{W_m}\sum_{r\in\mathcal R}w_r m_r
(\bar v_{g,r}-\bar v_{\cdot,r}).
\label{eq:rar_score}
\end{equation}
Here $\bar S=\frac{1}{G}\sum_{j=1}^{G}S_j$.
The same rubric mask is applied to every response, removing unreliable rubric
signals while retaining all responses. Training uses the gated score, with the
retained rubric contributions normalized by $W_m$.
Evaluation restores the full rubric set by setting $m_r=1$ for every
$r\in\mathcal R$.

\section{Experiments}
\label{sec:experiments}

\subsection{Experimental Setup}
\label{sec:setup}

\paragraph{Benchmarks.}
HealthBench provides conversation-specific physician rubrics
\citep{DBLP:journals/corr/abs-2505-08775}. We construct a category-balanced split of 4,000
training and 1,000 test conversations. PRBench comprises 1,100 expert-authored tasks
in finance and law \citep{DBLP:conf/acl/AkyurekGZGJGRMR26}. We split them evenly into
550 training and 550 test tasks and report results separately for the Non-Hard and
Hard subsets, each pooled across the two domains.
\onemillion contains 400 expert-curated tasks across five professional domains
\citep{DBLP:journals/corr/abs-2603-07980}. We use balanced training and test splits of
200 tasks each, with equal coverage of the five domains and the Global and CN
subsets.

\paragraph{Models and baselines.}
We evaluate Qwen3-4B-Instruct-2507 (Qwen3-4B) and Qwen3-30B-A3B-Instruct-2507
(Qwen3-30B-A3B) policies
\citep{DBLP:journals/corr/abs-2505-09388,
qwen2025qwen3_4b_instruct_2507,qwen2025qwen3_30b_a3b_instruct_2507}
with a gpt-oss-120b rubric judge
\citep{DBLP:journals/corr/abs-2508-10925}. All main experiments use high judge
reasoning effort. We additionally evaluate Qwen3-4B on HealthBench and PRBench
at low, medium, and high judge reasoning effort to measure sensitivity to this
setting. We compare \method with three baselines.
\rar~\citep{DBLP:journals/corr/abs-2507-17746} explicitly aggregates rubric
verdicts without a reliability gate,
\arl~\citep{DBLP:journals/corr/abs-2603-15646} optimizes one semantic rubric
class at a time, and \powthree~\citep{DBLP:journals/corr/abs-2605-20164}
emphasizes rubrics that exhibit contrast across rollouts.
All methods use $G=16$ rollouts per prompt.
For the rubric types in Figure~\ref{fig:reasoning_effort_agreement}, we
determine each rubric's discrimination and stability from its
between-response and within-response variance, respectively, as specified in
Appendix~\ref{app:classification_vs_gate}.
Appendix~\ref{app:implementation} details the experimental setup, and
Table~\ref{tab:setup} lists the complete optimization settings.
Appendix~\ref{app:threshold} analyzes the choice of the hyperparameter
$\tau$, and Appendix~\ref{app:gate_ablation} compares the reliability gate
with random rubric removal and a moment-based gate.

\begin{table}[t]
\caption{\textbf{Main results on held-out test sets (\%).} Entries are mean
$\pm$ standard deviation over three runs under high judge reasoning effort.
Avg.\ averages the five evaluation sets, $\Delta$ is the gain of \method over
\rar, and boldface marks the best result for each policy model. \rar averages
$K$ judgments per rubric and explicitly aggregates all rubrics.}
\label{tab:main_results}
\centering
\small
\renewcommand{\arraystretch}{1.12}
\begin{tabularx}{\linewidth}{@{}l*{6}{>{\centering\arraybackslash}X}@{}}
\toprule
\multirow{2}{*}{\textbf{Method}} &
\multirow{2}{*}{\textbf{HealthBench}} &
\multicolumn{2}{c}{\textbf{PRBench}} &
\multicolumn{2}{c}{\textbf{\onemillion}} &
\multirow{2}{*}{\textbf{Avg.}}\\
\cmidrule(lr){3-4}\cmidrule(lr){5-6}
& & Non-Hard & Hard & Global & CN & \\
\midrule
\policyband{7}{Qwen3-4B}\\
\rar              & 46.23\sd{0.61} & 39.56\sd{0.33} & 18.67\sd{0.35} & 23.57\sd{0.82} & 25.06\sd{0.79} & 30.62\\
\arl              & 47.17\sd{0.70} & 40.73\sd{0.30} & 19.22\sd{0.24} & 23.88\sd{0.69} & 25.49\sd{0.78} & 31.30\\
\powthree         & 47.66\sd{0.59} & 41.88\sd{0.20} & 19.88\sd{0.23} & 24.62\sd{0.51} & 26.44\sd{0.74} & 32.10\\
\oursrow
\textbf{\method}  & \textbf{49.71}\sd{0.28} & \textbf{43.01}\sd{0.23} & \textbf{20.76}\sd{0.13} & \textbf{26.11}\sd{0.29} & \textbf{28.96}\sd{0.66} & \textbf{33.71}\\
\oursrow
\gainrow & \gain{3.48} & \gain{3.45} & \gain{2.09} & \gain{2.54} & \gain{3.90} & \gain{3.09}\\
\midrule
\policyband{7}{Qwen3-30B-A3B}\\
\rar              & 69.74\sd{0.64} & 47.46\sd{0.74} & 38.70\sd{0.81} & 31.04\sd{0.78} & 32.30\sd{0.89} & 43.85\\
\arl              & 70.33\sd{0.99} & 48.73\sd{0.72} & 39.24\sd{0.77} & 31.31\sd{0.89} & 33.18\sd{0.83} & 44.56\\
\powthree         & 71.96\sd{0.69} & 50.26\sd{0.65} & 39.71\sd{0.59} & 32.28\sd{0.57} & 33.83\sd{0.74} & 45.61\\
\oursrow
\textbf{\method}  & \textbf{74.34}\sd{0.58} & \textbf{52.23}\sd{0.33} & \textbf{41.67}\sd{0.41} & \textbf{33.83}\sd{0.68} & \textbf{35.47}\sd{0.58} & \textbf{47.51}\\
\oursrow
\gainrow & \gain{4.60} & \gain{4.77} & \gain{2.97} & \gain{2.79} & \gain{3.17} & \gain{3.66}\\
\bottomrule
\end{tabularx}
\end{table}

\subsection{Results}
\label{sec:results}

We summarize the results as seven findings, as shown below.

\finding{\method outperforms all three baselines on every benchmark.}
As shown in Table~\ref{tab:main_results}, \method achieves the best
performance on HealthBench, PRBench, and \onemillion. Relative to \rar,
\method improves Qwen3-4B and Qwen3-30B-A3B by
$3.09\%$ and $3.66\%$ on average, respectively. Notably,
Qwen3-4B improves by
$3.90\%$ on the
\onemillion CN subset. The larger
Qwen3-30B-A3B policy benefits more from \method. On HealthBench, for example,
its improvement over \rar is
$4.60\%$, compared with $3.48\%$ for Qwen3-4B. Beyond the mean gains,
\method also has the smallest run-to-run standard deviation in eight of the
ten comparisons. These results are consistent with our central hypothesis
that retaining stable, discriminative rubrics concentrates optimization on
reliable response-level contrasts, thereby improving the fidelity and
discriminative power of the learning signal.
Appendix~\ref{app:detailed_results} further stratifies the results by domain
and difficulty level.

\begin{figure*}[t]
\centering
\includegraphics[width=\textwidth]{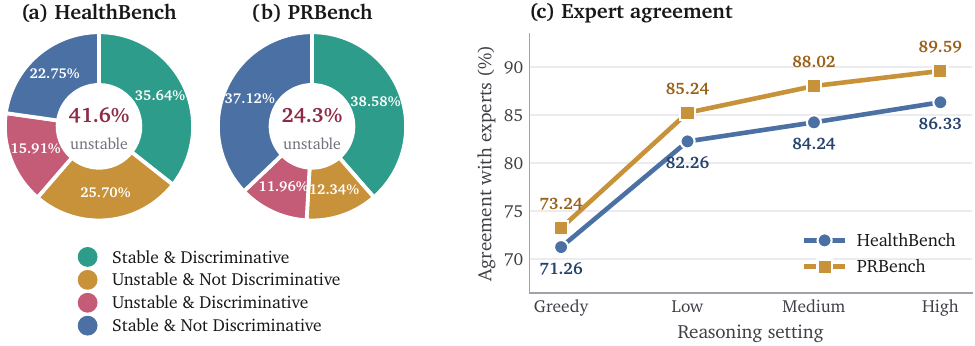}
\caption{\textbf{Rubric dynamics and expert agreement across reasoning
settings.} \textbf{(a)} Distribution of rubric types during
Qwen3-4B training on HealthBench, evaluated by the judge model at high
reasoning effort. \textbf{(b)} Distribution of rubric types during training
on PRBench under the same setting. Rubric types are defined by judgment
stability and within-group response discrimination, and each donut center gives
the combined share of unstable rubrics. \textbf{(c)} Agreement
(\%) between the judge model and expert annotations under greedy
decoding and low, medium, and high reasoning effort.}
\label{fig:reasoning_effort_agreement}
\end{figure*}

\finding{Unstable rubrics make up a large share of the training signal,
more than 40\% on HealthBench.}
Figures~\ref{fig:reasoning_effort_agreement}(a)
and~\ref{fig:reasoning_effort_agreement}(b) show that all four combinations of
stability and discrimination occur at non-negligible rates during training,
underscoring that these two properties capture distinct aspects of rubric
quality. Unstable rubrics are prevalent on both benchmarks and account for
more than $40\%$ of the learning signals on HealthBench. Evaluator
inconsistency is therefore not a marginal failure mode and, without explicit
control, can exert substantial spurious optimization pressure. \method
provides such explicit control by filtering out unstable rubric signals,
thereby grounding policy updates in response-level differences that are
reproducible across repeated judgments. Consistent with this mechanism, its
gains over the strongest baseline are larger on HealthBench, where unstable
rubrics are more prevalent than on PRBench.

\finding{Greedy decoding improves judgment stability but lowers agreement
with experts by more than 15\%.}
Figure~\ref{fig:reasoning_effort_agreement}(c) compares the judge model's judgments
with expert annotations across reasoning-effort settings. On both benchmarks,
agreement increases with configured reasoning effort. Greedy decoding can
improve stability by removing sampling randomness. However, it lowers
expert agreement by more than $15\%$ relative to high effort on both
benchmarks, so the judge model makes substantially more incorrect judgments.
These results suggest a trade-off between expert alignment and judgment
stability. \method addresses this trade-off by
retaining high-effort judgments for their stronger observed expert alignment
while explicitly identifying and filtering unstable rubric signals.

\begin{figure*}[t]
\centering
\includegraphics[width=\textwidth]{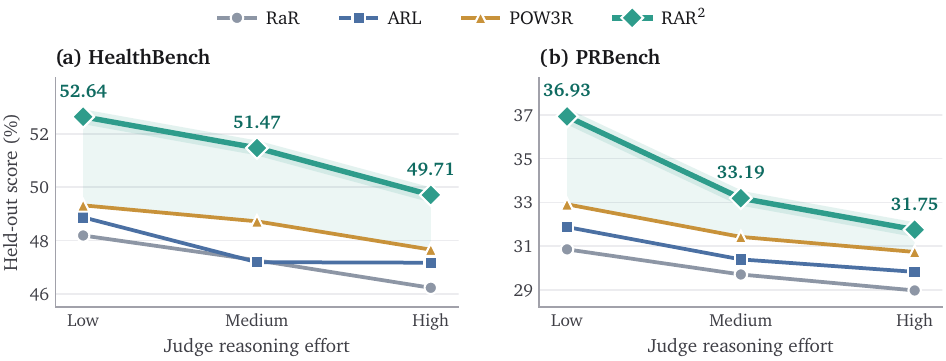}
\caption{\textbf{Method performance across the judge model's reasoning efforts.}
Held-out Qwen3-4B scores (\%) at low, medium, and high judge
reasoning effort on \textbf{(a)} HealthBench and \textbf{(b)} PRBench.}
\label{fig:reasoning_effort_performance}
\vspace{10pt}
\includegraphics[width=\textwidth]{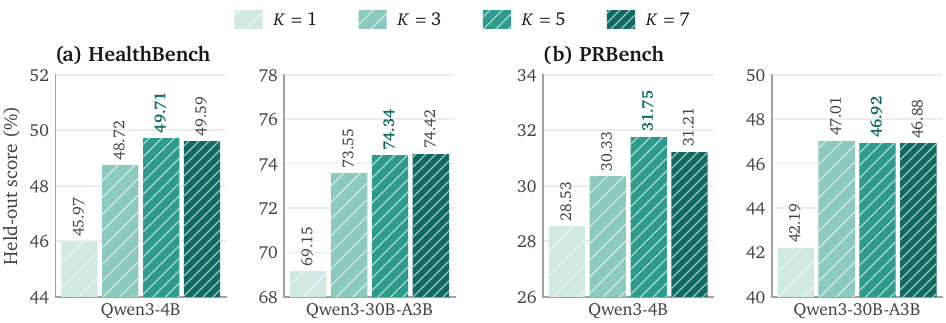}
\caption{\textbf{Effect of the number of training judgments $K$.} Model
performance (\%) of Qwen3-4B and Qwen3-30B-A3B trained with \method on
\textbf{(a)} HealthBench and \textbf{(b)} PRBench.}
\label{fig:judgment_count_performance}
\end{figure*}

\finding{\method yields the largest gains when the judge is weakest, at
low reasoning effort.}
Figure~\ref{fig:reasoning_effort_performance} shows that \method consistently
outperforms all baselines under every reasoning-effort setting of the judge
model. The gains are most pronounced at low effort, where the relative
improvement over the strongest baseline, \powthree, reaches $6.73\%$ on
HealthBench and $12.22\%$ on PRBench. This may be attributed to the
larger share of unstable and non-discriminative signals produced by the
low-effort judge. By filtering them out, \method focuses policy updates on
stable and discriminative rubric signals and thereby substantially improves
training.

\finding{Performance improves with the number of judgments per rubric but
saturates at $K=5$.}
We examine how the number of training judgments $K$ affects final
performance, as illustrated in Figure~\ref{fig:judgment_count_performance}. As
shown in Eq.~\ref{eq:single_judgment_likelihood}, a single judgment is
uninformative about $\phi$, so \method reduces to \rar at $K=1$. Once $K\ge2$
renders $\phi$ identifiable, performance improves substantially but saturates
by $K=5$. Increasing $K$ from $5$ to $7$ incurs $40\%$ more judge calls yet
changes performance by at most $0.54\%$, even degrading it in three of the four
settings. As indicated by
the Spearman--Brown formula in Eq.~\ref{eq:rubric_reliability}, the marginal
reliability gain decays as $O(K^{-2})$, whereas judge cost grows linearly. We
therefore adopt $K=5$ to balance performance and judge cost.

\begin{figure*}[t]
\centering
\includegraphics[width=\textwidth]{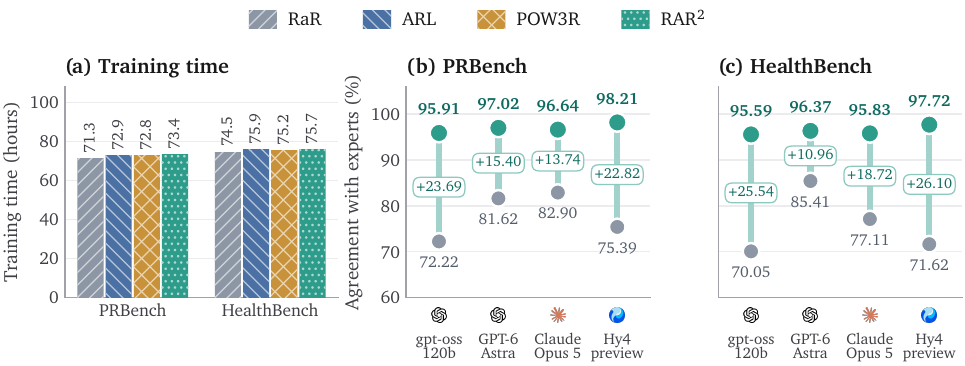}
\caption{\textbf{Efficiency and cross-judge agreement with expert
assessments.} \textbf{(a)} Training time for 1,000 steps with batch size 16.
\textbf{(b)} and \textbf{(c)} Agreement of \rar and \method with expert assessments under
four rubric judges, with the gain of \method over \rar marked on each
connector. For each benchmark, both methods and all judges are
evaluated on the same 100 samples drawn during training. Expert assessments
are provided by three human experts.}
\label{fig:efficiency_cross_judge_agreement}
\end{figure*}

\finding{By exploiting the mutual independence of judge calls, \method adds
less than 3\% to training time over \rar.}
Figure~\ref{fig:efficiency_cross_judge_agreement}(a) compares the wall-clock
training time of different methods under the fixed 1,000-step protocol.
Relative to \rar, \method incurs only a 2.9\% training-time overhead on PRBench
and a 1.6\% overhead on HealthBench. Its runtime is comparable to those of
\arl and \powthree, within 1\% of each on both benchmarks, and is even slightly
shorter than that of \arl on HealthBench. This modest overhead stems from the
mutual independence of judge calls, which enables their concurrent execution.
Consequently, \method can estimate rubric reliability without a substantial
increase in wall-clock time.

\finding{The training signal of \method reaches more than 95\% agreement
with human experts.}
Figures~\ref{fig:efficiency_cross_judge_agreement}(b)
and~\ref{fig:efficiency_cross_judge_agreement}(c) compare the agreement
between expert assessments and the learning signals produced by various judge
models during training. \method improves agreement for every judge--benchmark
pair, with gains over \rar ranging from $13.74\%$ to $23.69\%$ on PRBench and
from $10.96\%$ to $26.10\%$ on HealthBench. Hy4 preview exhibits the largest
improvement, with an agreement gain of $26.10\%$ on HealthBench. Agreement with
\method remains above $95\%$ across all eight settings, showing consistent
gains across judge models.
Agreement is computed on the same sampled rubric--response decisions for
both methods. Appendix~\ref{app:further_analysis} gives the annotation
protocol and further analyses.

\section{Related Work}
\label{sec:related}

\paragraph{Rubric-based evaluation.}
Rubric-based LLM evaluation decomposes open-ended quality into explicit,
task-specific criteria, making judgments more interpretable and actionable
\citep{DBLP:conf/iclr/KimS0JLLYSKTS24,DBLP:conf/acl/HashemiERDK24}. Expert-authored rubrics provide
domain-specific evaluation objectives in medicine and professional reasoning
\citep{DBLP:journals/corr/abs-2505-08775,DBLP:conf/acl/AkyurekGZGJGRMR26}. Related frameworks
operationalize criteria through generated evaluation steps, instance-relevant
skills, self-adaptive question-level rubrics, and open evaluator models
\citep{DBLP:conf/emnlp/LiuIXWXZ23,DBLP:conf/iclr/YeKKHKJTKS24,
DBLP:conf/emnlp/FanWWZ24,DBLP:conf/emnlp/KimSLLSWNL0S24}. RRD uses recursive
decomposition and filtering to expand rubric coverage and remove misaligned
or redundant criteria \citep{DBLP:journals/corr/abs-2602-05125}. Autorubric consolidates
heterogeneous rubric types, judge ensembles, and psychometric diagnostics
within a unified evaluation framework \citep{Rao2026AutorubricAU}. Rubrics as
Rewards (\rar) further converts instance-specific rubric verdicts into
structured signals for on-policy reinforcement learning
\citep{DBLP:journals/corr/abs-2507-17746}.

\paragraph{Reliability and uncertainty of LLM feedback.}
LLM judgments can exhibit substantial run-to-run variability
\citep{DBLP:journals/corr/abs-2412-12509,
DBLP:journals/corr/abs-2405-01724,DBLP:conf/coling/LeeHT25,
DBLP:journals/corr/abs-2606-13685}. Complementary work quantifies the
confidence of aggregate rankings \citep{DBLP:conf/eacl/StureborgAS24},
constructs conformal prediction intervals around judge scores
\citep{DBLP:conf/emnlp/ShengLHZK25}, or pools verdicts from a panel of
diverse judges \citep{DBLP:journals/corr/abs-2404-18796}. Classical intraclass-correlation models
quantify stable between-target variation relative to measurement error, while
the Spearman--Brown formula predicts the reliability gain from averaging
multiple measurements \citep{shrout1979icc,spearman1910correlation}.
For each rubric, \method estimates the posterior probability that stable
differences among current-policy responses outweigh the judge noise, and
drops rubrics that fall below a confidence threshold.

\paragraph{Decomposed and adaptive rewards for RL.}
Fine-grained RLHF \citep{DBLP:conf/nips/WuHSDSASOH23} and LLM-graded rule-based
safety rewards \citep{DBLP:conf/nips/MuHHAVKLBSW24} both provide decomposed
training signals. Multi-objective alignment trades off
diverse rewards through reward-specific model-weight interpolation
\citep{DBLP:conf/nips/RameCDGSSC23} or objective-weighted DPO
\citep{DBLP:conf/acl/ZhouLS00O024}. Reward-model overoptimization has been
characterized in synthetic settings with a proxy and a gold reward model
\citep{DBLP:conf/icml/GaoSH23}, and conservative reward-model ensembles
mitigate it in related settings \citep{DBLP:conf/iclr/CosteAK024}. Existing
methods adapt policy optimization according to observed rubric outcomes.
POW3R emphasizes rubrics exhibiting rollout contrast
\citep{DBLP:journals/corr/abs-2605-20164}, Focal Reward prioritizes unsaturated
criteria \citep{DBLP:journals/corr/abs-2605-26579}, and ARL optimizes semantic
rubric classes sequentially \citep{DBLP:journals/corr/abs-2603-15646}. Rubric
Dropout instead randomizes group-shared criteria to reduce proxy exploitation
\citep{DBLP:journals/corr/abs-2608-11669}. These approaches differ in how they
select or weight rubrics, yet all of them overlook judgment stability and thus
cannot distinguish genuine response differences from contrast induced by judge
noise.

\section{Conclusion}
\label{sec:conclusion}

This work studies rubric-based rewards for reinforcement learning and
identifies judgment stability and response discrimination as two complementary
requirements. Because rubric contrasts from a single evaluation can
conflate genuine response differences with judge inconsistency, we introduce
\method, which uses multiple rubric-level verdicts and a Beta--Binomial
posterior to estimate each rubric's reliability and drops the unreliable
ones. \method therefore grounds policy updates in reproducible response
differences rather than judge noise.
Across three benchmarks and two policy scales, \method consistently
outperforms strong baselines and improves agreement between the training
signal and expert assessments.
These results demonstrate that explicitly controlling reward reliability
improves both the fidelity and effectiveness of rubric-guided learning.

\subsubsection*{Ethics statement}
Our experiments use three publicly released benchmarks. HealthBench
\citep{DBLP:journals/corr/abs-2505-08775} is distributed under the MIT License,
PRBench \citep{DBLP:conf/acl/AkyurekGZGJGRMR26} under CC~BY~4.0, and
\onemillion \citep{DBLP:journals/corr/abs-2603-07980} under the Apache
License~2.0. Our use of these benchmarks, including re-splitting them into
training and test sets, complies with their licenses. The human experts who
provided the expert assessments were recruited through legitimate channels and
compensated at an hourly rate commensurate with the local average wage for data
annotation.

\fullauthorlist
\vspace{1.2\baselineskip}

\bibliographystyle{plainnat}
\bibliography{references}

\FloatBarrier
\appendix
\section*{Appendix}
\section{Statistical Derivations}
\label{app:derivations}

\subsection{Derivation of Eq.~\ref{eq:rubric_reliability}}

For $q_g\sim\mathrm{Beta}(\alpha,\beta)$ with
$\alpha=\mu\kappa$, $\beta=(1-\mu)\kappa$, and
$\kappa=\phi^{-1}-1$,
\begin{equation}
    \E[q_g]=\mu,\qquad
    \Var(q_g)=\frac{\mu(1-\mu)}{\kappa+1}
    =\mu(1-\mu)\phi.
\end{equation}
Using $\E[q_g(1-q_g)]=\E[q_g]-\E[q_g^2]$ and
$\E[q_g^2]=\Var(q_g)+\mu^2$,
\begin{equation}
    \E[q_g(1-q_g)]
    =\mu(1-\mu)-\mu(1-\mu)\phi
    =\mu(1-\mu)(1-\phi).
\end{equation}
Conditional on $q_g$,
\begin{equation}
    \E[\bar v_g\mid q_g]=q_g,\qquad
    \Var(\bar v_g\mid q_g)
    =\frac{q_g(1-q_g)}{K}.
\end{equation}
The law of total variance therefore gives
\begin{align}
    \Var(\bar v_g)
    &=
    \Var(q_g)
    +\frac{1}{K}\E[q_g(1-q_g)]
    \label{eq:mean_verdict_variance}\\
    &=
    \mu(1-\mu)
    \left[\phi+\frac{1-\phi}{K}\right].
\end{align}
Dividing the latent between-response component by the total yields
\begin{equation}
    \frac{\phi}{\phi+(1-\phi)/K}
    =\frac{K\phi}{1+(K-1)\phi}
    =\rho_K.
\end{equation}
For two distinct judgments $k\neq k'$ of the same response, conditional
independence and the law of total covariance give
$\Cov(v_{gk},v_{gk'})=\Var(q_g)=\mu(1-\mu)\phi$. Each verdict is marginally
$\mathrm{Bernoulli}(\mu)$, so $\Var(v_{gk})=\mu(1-\mu)$ and
$\Corr(v_{gk},v_{gk'})=\phi$. Thus $\phi$ is the single-verdict intraclass
correlation, and $\rho_K$ is the Spearman--Brown reliability of the mean of
$K$ exchangeable verdicts.

\subsection{Moment decomposition}
\label{app:moment}

Fix one rubric, omit its index $r$, and write
$\bar v_{\cdot}:=\frac{1}{G}\sum_{g=1}^{G}\bar v_g$. For $G\ge2$, $K\ge2$, and
complete equal-$K$ rows, the diagnostic moment estimator is
\begin{align}
\widehat\sigma_W^2
&=\frac{1}{G}\sum_{g=1}^{G}
\frac{n_g(K-n_g)}{K(K-1)},\\
s_{\bar v}^{2}
&=\frac{1}{G-1}\sum_{g=1}^{G}
(\bar v_g-\bar v_{\cdot})^2,\\
\widehat\sigma_B^2
&=\max\!\left\{s_{\bar v}^{2}-\widehat\sigma_W^2/K,\,0\right\},\\
\widehat\rho_{\mathrm{mom}}
&=\frac{\widehat\sigma_B^2}
{\widehat\sigma_B^2+\widehat\sigma_W^2/K}.
\end{align}
Here $\widehat\sigma_W^2$ is the average within-response judge variance,
$s_{\bar v}^{2}$ is the observed variance across response means,
$\widehat\sigma_B^2$ estimates latent between-response variance, and
$\widehat\rho_{\mathrm{mom}}$ is the corresponding estimate of $\rho_K$.
Because $n_g\mid q_g\sim\mathrm{Binomial}(K,q_g)$, we have
$\E[n_g(K-n_g)\mid q_g]=K(K-1)\,q_g(1-q_g)$, so $\widehat\sigma_W^2$ is unbiased
for $\sigma_W^2=\E[q_g(1-q_g)]$. By Eq.~\ref{eq:mean_verdict_variance}, the
i.i.d.\ response means $\bar v_g$ have variance $\sigma_B^2+\sigma_W^2/K$ with
$\sigma_B^2=\Var(q_g)$. Before truncation, $s_{\bar v}^{2}-\widehat\sigma_W^2/K$
is therefore unbiased for $\sigma_B^2$. Whenever $s_{\bar v}^{2}>0$, the
estimator reduces to
$\widehat\rho_{\mathrm{mom}}=\max\{1-\widehat\sigma_W^2/(Ks_{\bar v}^{2}),\,0\}$,
which is the one-way $\mathrm{ICC}(1,K)$ of \citet{shrout1979icc} truncated at
zero.

When the judge returns the same verdict for every response and every
repetition, both $\widehat\sigma_W^2$ and $s_{\bar v}^{2}$ are zero, so the
defining ratio is undefined and we set $\widehat\rho_{\mathrm{mom}}=0$. With
small $G$ and low $\phi$, sampling noise often pushes
$s_{\bar v}^{2}-\widehat\sigma_W^2/K$ below zero, and truncation then reports
zero reliability even for rubrics with genuine between-response differences.
For these reasons, the training gate of \method relies on the posterior
probability $R_r$ in Eq.~\ref{eq:posterior_summaries} and the fail-open
safeguards in Eq.~\ref{eq:gate}, while $\widehat\rho_{\mathrm{mom}}$ serves
only as a diagnostic.

\subsection{Properties of the reliability gate}
\label{app:gate_geometry}

Substituting $\sigma_B^2=\mu(1-\mu)\phi$ and $\sigma_W^2=\mu(1-\mu)(1-\phi)$
into $\rho_K=\sigma_B^2/(\sigma_B^2+\sigma_W^2/K)$ gives, for $K\ge1$,
\begin{equation}
\rho_K\ge\tfrac12
\iff
\sigma_B^2\ge\frac{\sigma_W^2}{K}
\iff
\phi\ge\frac{1}{K+1}.
\label{eq:gate_equivalence}
\end{equation}
The gate statistic therefore has three equivalent readings,
\begin{equation}
R_r
=\Prob\!\left(\rho_K\ge\tfrac12\;\middle|\;\mathbf n_r\right)
=\Prob\!\left(\sigma_B^2\ge\frac{\sigma_W^2}{K}\;\middle|\;\mathbf n_r\right)
=\Prob\!\left(\phi\ge\frac{1}{K+1}\;\middle|\;\mathbf n_r\right).
\label{eq:gate_three_readings}
\end{equation}
A rubric is retained when the posterior places at least mass $\tau$ on
between-response contrast that exceeds the judge noise remaining after
averaging $K$ verdicts. Stability and discrimination thus enter one
inequality that trades them off: a small contrast suffices when judgments are
reproducible, and noisier judgments require a larger contrast. Because the
right-hand form of Eq.~\ref{eq:gate_equivalence} does not involve $\mu$, the
gated event does not depend on a rubric's pass rate, so rubrics that flag rare
but consistently judged failures are not penalized.

\section{Implementation Details}
\label{app:implementation}

\subsection{Experimental configuration}

\paragraph{Hardware and judges.}
We train Qwen3-4B on 8 NVIDIA H20 GPUs and Qwen3-30B-A3B on 32 H20 GPUs. The
rubric judge gpt-oss-120b is served on a separate judge server with 16 H20
GPUs. The cross-judge analysis in
Figure~\ref{fig:efficiency_cross_judge_agreement} additionally uses GPT-6
Astra \citep{openai2026gpt6astra}, Claude Opus 5
\citep{anthropic2026claudeopus5}, and Hy4 preview
\citep{tencent2026hy4preview} as judges.

\paragraph{Training reward.}
Under \method, each training response $y_g$ receives the reliability-gated
score $S_g$ in Eq.~\ref{eq:rar_score}. The training reward further includes a
response-length penalty and a format reward with weight $0.1$. Advantages
follow GRPO \citep{DBLP:journals/corr/abs-2402-03300}, which centers and
standardizes the rewards within each group of $G$ responses.
Table~\ref{tab:setup} lists the remaining hyperparameters.

\begin{table}[ht]
\caption{\textbf{Training and evaluation settings.} The gate threshold and the
protected weight apply only to \method.}
\label{tab:setup}
\centering
\small
\begin{tabular}{lccc}
\toprule
Setting & HealthBench & PRBench & \onemillion\\
\midrule
Train / test prompts & 4,000 / 1,000 & 550 / 550 & 200 / 200\\
Learning rate & $1{\times}10^{-6}$ & $5{\times}10^{-6}$ &
$5{\times}10^{-6}$\\
Warmup ratio & 0 & 0.1 & 0.1\\
Training epochs & 8 & 12 & 12\\
\midrule
Prompt batch size & \multicolumn{3}{c}{16}\\
Responses per prompt $G$ & \multicolumn{3}{c}{16}\\
Judgments per response $K$, train / test & \multicolumn{3}{c}{5 / 3}\\
Rollout temperature & \multicolumn{3}{c}{1.0}\\
Gate threshold $\tau$ & \multicolumn{3}{c}{0.55}\\
Protected weight & \multicolumn{3}{c}{$|w_r|\ge8$}\\
\bottomrule
\end{tabular}
\end{table}

\subsection{Evaluation metric and protocol}
\label{app:eval_metric}
\label{app:eval_protocol}

We report the Expert Score on all benchmarks, following the rubric scoring rule
of HealthBench \citep{DBLP:journals/corr/abs-2505-08775} and PRBench
\citep{DBLP:conf/acl/AkyurekGZGJGRMR26}.
For the $j$-th test prompt $x_j$, $j=1,\ldots,N$, let $y_j$ be the policy
response and let $c_r(x_j,y_j)$ be the verdict on rubric $r$ averaged over the
test judgments. The prompt score $s_j$ is $S^{\mathrm{RaR}}$ in
Eq.~\ref{eq:rar_baseline_score} computed on the full rubric set, that is, the
weighted sum of rubric verdicts divided by the total positive weight $W$. The
benchmark score, reported in percent, is
\begin{equation}
    \mathrm{Score}
    =\max\Bigl\{0,\ \frac{1}{N}\sum_{j=1}^{N}s_j\Bigr\}.
\end{equation}
The length penalty and the format reward are excluded from evaluation.
All held-out results use one fixed protocol, regardless of the method or
the training judge: the evaluation judge is gpt-oss-120b at high reasoning
effort, each rubric is judged $K=3$ times, and every method is scored on the
complete rubric set with the gate disabled. In
Figure~\ref{fig:reasoning_effort_performance}, only the reasoning effort of
the training judge varies. The expert agreement in
Figures~\ref{fig:reasoning_effort_agreement}(c),
\ref{fig:efficiency_cross_judge_agreement}(b),
and~\ref{fig:efficiency_cross_judge_agreement}(c) is instead measured on
training-time rubric decisions.

\subsection{Distributed implementation}

\paragraph{Row aggregation.}
During training, the $G$ responses to a prompt are scored by several reward
workers in parallel. Each response yields one row of $K$ verdicts per rubric,
and the rows of a group arrive asynchronously in different processes.
Algorithm~\ref{alg:rar_full} shows how these rows are combined into a single
rubric mask per prompt group. A coordinator stores rows under a stable prompt key and a round
counter. A per-prompt asynchronous lock makes reading the round, appending a
row, and finalizing the mask atomic, while judge calls remain outside the
lock. The coordinator waits for all $G$ rows up to a fixed time limit. Once
the limit is reached, it stops waiting and computes the mask from the rows
that have arrived. If fewer than three rows are available, the gate is skipped
and every rubric is retained. The committed mask is cached, so late rows are
scored with the same mask as the rest of their group.

\begin{algorithm}[!htbp]
\caption{Distributed reliability-gated rubric reward}
\label{alg:rar_full}
\begin{algorithmic}[1]
\Require prompt key $u$, prompt $x$, rubric set
$\mathcal R=\{r_1,\ldots,r_M\}$ with weights $w_r$,
expected group size $G$, judgments per response $K$, threshold $\tau$
\State sample responses $y_1,\ldots,y_G$
\ForAll{responses $y_g$ in parallel}
    \State request $K$ rubric-verdict vectors from the fixed judge protocol
    \State submit valid vectors under prompt key $u$ and its current round
\EndFor
\State coordinator waits for all $G$ rows or until the time limit
\If{fewer than three rows are available}
    \State return $m_r=1$ for every $r\in\mathcal R$
\Else
    \ForAll{$r\in\mathcal R$}
        \State form $\mathbf n$ for rubric $r$ from the complete rows
        \State infer $R_r$
        \If{rubric $r$ is protected or has insufficient evidence}
            \State $m_r\gets1$
        \Else
            \State $m_r\gets\ind[R_r\ge\tau]$
        \EndIf
    \EndFor
\EndIf
\If{$\sum_{r\in\mathcal R}m_r\max(w_r,0)=0$}
    \State $m_r\gets1$ for every $r\in\mathcal R$
\EndIf
\State cache the shared rubric mask for late rows and advance the round
\State score each response with the same $(m_r)_{r\in\mathcal R}$
\State center and standardize scores within the group; update the policy
\end{algorithmic}
\end{algorithm}

\paragraph{Posterior evaluation.}
Only complete rows, in which all $K$ judgments are valid, enter the count
vector $\mathbf n$ in Eq.~\ref{eq:rubric_counts}. The posterior in
Eq.~\ref{eq:posterior} is evaluated on a $120\times120$ grid of equally spaced
values of $(\mu,\phi)$ in $[0.005,0.995]^2$. Beta--Binomial log-probabilities
are precomputed for every count and grid cell, so each rubric requires only a
histogram of its counts, a weighted sum of cached terms, and a log-sum-exp
normalization. The gate statistic $R_r$ is the posterior mass of the grid cells
with $\rho_K\ge\tfrac12$, which is equivalent to $\phi\ge1/(K+1)$.

\setlength{\intextsep}{6pt plus 2pt minus 2pt}
\captionsetup{skip=5pt}

\section{Further Analysis}
\label{app:further_analysis}

\subsection{Effect of the gate threshold \texorpdfstring{$\tau$}{tau}}
\label{app:threshold}

\begin{figure}[!htbp]
\centering
\includegraphics[width=0.63\linewidth]{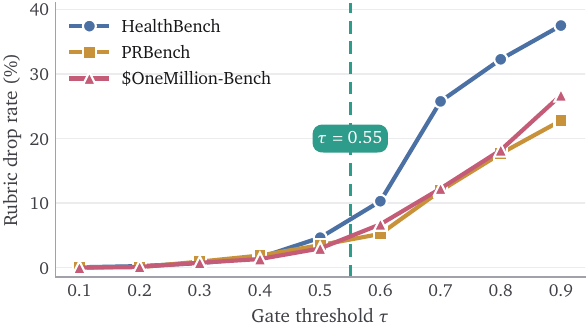}
\caption{\textbf{Rubric drop rate under different gate thresholds $\tau$.}
The drop rate is the fraction of rubrics removed by the reliability gate
during training. The dashed line marks the default $\tau=0.55$.}
\label{fig:threshold_drop_rate}
\end{figure}

Figure~\ref{fig:threshold_drop_rate} compares the rubric drop rate during
training under different thresholds $\tau$, where the drop rate is the
fraction of rubrics removed by the reliability gate in Eq.~\ref{eq:gate}.
For small $\tau$, the
gate retains almost all rubrics and \method effectively reduces to \rar. At
$\tau=0.1$, the drop rate is at most $0.05\%$ on every benchmark. The drop
rate rises sharply on all three benchmarks once $\tau$ reaches $0.7$ and
continues to increase for larger $\tau$. Such aggressive filtering removes a
substantial share of the learning signal available for training. We therefore
set $\tau=0.55$, below the point at which the drop rate rises sharply.

To measure how $\tau$ affects the trained policy,
Table~\ref{tab:threshold_performance} compares Qwen3-4B trained at
$\tau=0.40$, $\tau=0.55$, and $\tau=0.70$ with all other settings unchanged.
At $\tau=0.40$, the gate removes only $1.57\%$, $1.87\%$, and $1.33\%$ of
rubrics on HealthBench, PRBench, and \onemillion, so unstable learning
signals are left largely unfiltered and \method gains just $0.18$ to $0.56$
percentage points over \rar. At $\tau=0.70$, the removal rates rise to $25.76\%$,
$11.91\%$, and $12.25\%$, and performance degrades on every slice to below
\rar because useful rubrics are discarded along with noisy ones. We
therefore set $\tau=0.55$, which balances the reliability of the learning
signal against the amount of signal.

\begin{table}[H]
\caption{\textbf{Held-out Qwen3-4B scores (\%) under different gate
thresholds $\tau$.} Boldface marks the best score in each column.}
\label{tab:threshold_performance}
\centering
\footnotesize
\setlength{\tabcolsep}{4pt}
\begin{tabular}{@{}lccccc@{}}
\toprule
\multirow{2}{*}{Gate setting} &
\multirow{2}{*}{HealthBench} &
\multicolumn{2}{c}{PRBench} &
\multicolumn{2}{c}{\onemillion}\\
\cmidrule(lr){3-4}\cmidrule(l){5-6}
& & Non-Hard & Hard & Global & CN\\
\midrule
\rar, no gate            & 46.23 & 39.56 & 18.67 & 23.57 & 25.06\\
$\tau=0.40$              & 46.71 & 40.12 & 18.85 & 23.98 & 25.34\\
$\tau=0.55$              & \textbf{49.71} & \textbf{43.01} & \textbf{20.76} &
                           \textbf{26.11} & \textbf{28.96}\\
$\tau=0.70$              & 44.62 & 39.54 & 18.49 & 23.11 & 24.79\\
\bottomrule
\end{tabular}
\end{table}

\subsection{Ablations on the reliability gate}
\label{app:gate_ablation}

We compare \method with two gate variants that differ only in how the mask
$(m_r)_{r\in\mathcal R}$ is
produced. \emph{Random removal} fixes a random mask for each prompt before
training, with per-benchmark removal rates matched to those of \method at
$\tau=0.55$, and renormalizes the retained rubrics by $W_m$ as in
Eq.~\ref{eq:rar_score}. The \emph{moment-based gate} keeps the reliability
criterion but replaces the posterior with the point estimate
$\widehat\rho_{\mathrm{mom}}$ of Appendix~\ref{app:moment}, under the same
fail-open safeguards. Its threshold $\tfrac12$ is the boundary of the
posterior event of \method, at which stable between-response differences
account for half of the variance in mean verdicts. Outside the safeguards,
the two gates are
\[
m_r^{\mathrm{mom}}
=\ind\!\left[\widehat\rho_{\mathrm{mom},r}\ge\tfrac12\right]
\qquad\text{and}\qquad
m_r
=\ind\!\left[\Prob\!\left(\rho_K\ge\tfrac12\;\middle|\;\mathbf n_r\right)
\ge\tau\right],
\]
where $\tau=0.55$ is a required posterior probability rather than a second
reliability level. The two gates thus target the same boundary and differ
only in how they treat estimation uncertainty.

\begin{table}[H]
\caption{\textbf{Gate ablations with Qwen3-4B (\%).} Entries are mean
$\pm$ standard deviation over three runs with seeds 41, 42, and 43. Boldface
marks the best mean in each column.}
\label{tab:gate_ablation}
\centering
\footnotesize
\setlength{\tabcolsep}{2pt}
\begin{tabular}{@{}lccccc@{}}
\toprule
\multirow{2}{*}{Method} &
\multirow{2}{*}{HealthBench} &
\multicolumn{2}{c}{PRBench} &
\multicolumn{2}{c}{\onemillion}\\
\cmidrule(lr){3-4}\cmidrule(l){5-6}
& & Non-Hard & Hard & Global & CN\\
\midrule
\rar              & 46.23\sd{0.6106} & 39.56\sd{0.3300} & 18.67\sd{0.3510} &
                    23.57\sd{0.8217} & 25.06\sd{0.7901}\\
Random removal    & 43.09\sd{0.5264} & 38.85\sd{0.3527} & 12.26\sd{0.6925} &
                    19.33\sd{0.4192} & 22.72\sd{0.5734}\\
Moment-based gate & 48.65\sd{0.4287} & 42.57\sd{0.3385} & 20.36\sd{0.5295} &
                    25.91\sd{0.3476} & 28.45\sd{0.4926}\\
\method           & \textbf{49.71}\sd{0.2825} & \textbf{43.01}\sd{0.2277} &
                    \textbf{20.76}\sd{0.1257} & \textbf{26.11}\sd{0.2884} &
                    \textbf{28.96}\sd{0.6582}\\
\bottomrule
\end{tabular}
\end{table}

As illustrated in Table~\ref{tab:gate_ablation}, random removal falls below
\rar on all five slices, so removal alone is not beneficial. The
moment-based gate improves on \rar by $1.69$ to $3.39$ percentage points,
showing that
the reliability criterion is what makes removal useful. \method attains a
higher mean than the moment-based gate on every slice, by $0.20$ to $1.06$
percentage points, and the gap exceeds the standard deviations of both
methods on
HealthBench and PRBench Non-Hard. With $G=16$ and $K=5$, the point estimate
$\widehat\rho_{\mathrm{mom}}$ is noisy and often truncated to zero, whereas
the posterior accounts for this estimation uncertainty.

\subsection{Rubric type classification}
\label{app:classification_vs_gate}

The rubric types in Figures~\ref{fig:reasoning_effort_agreement}(a)
and~\ref{fig:reasoning_effort_agreement}(b) are assigned per prompt group by
thresholding the two empirical variance components of
Appendix~\ref{app:moment} separately:
\begin{equation}
\text{discriminative}
\iff
s_{\bar v}^{2}\ge\delta_B,
\qquad
\text{stable}
\iff
\widehat\sigma_W^{2}\le\delta_W,
\label{eq:quadrant_rule}
\end{equation}
with $\delta_B=0.01$ and $\delta_W=0.05$. The first rule requires the
response means to spread by at least $0.1$ in standard deviation. At $K=5$,
the second allows judge disagreement on at most about a quarter of the
responses. These descriptive rules measure how often each phenomenon occurs
and differ from the training gate, which thresholds the posterior
probability of the ratio event in Eq.~\ref{eq:gate_three_readings} and
applies fail-open safeguards. A noisy rubric still passes the gate when its
between-response contrast is large, and a rubric with uninformative counts is
retained because the gate removes rubrics only on evidence of unreliability.

\subsection{Expert agreement: protocol and confusion matrices}
\label{app:expert_protocol}

Expert agreement in Figures~\ref{fig:reasoning_effort_agreement}(c),
\ref{fig:efficiency_cross_judge_agreement}(b),
and~\ref{fig:efficiency_cross_judge_agreement}(c) is computed over items,
where an item is one rubric criterion applied to one policy response. For each
benchmark, we uniformly sample $100$ training examples from the first epoch
and annotate all items they induce, using the same items for both methods and
all four judges. Two experts label each item independently, with Pass or Fail
for \rar and with Pass, Fail, or Drop for \method, where Drop means that the
criterion should not contribute to the reward for that response. A third
expert adjudicates disagreements. Agreement is the fraction of all sampled
items on which a method's decision matches the final expert label. Items that
\method drops remain in the denominator and count as agreement only when the
experts also chose Drop. Tables~\ref{tab:confusion_rar} and~\ref{tab:confusion_rar2}
give the confusion matrices behind
Figure~\ref{fig:efficiency_cross_judge_agreement}(b).

\begin{table}[H]
\caption{\textbf{\rar decisions against final expert labels on PRBench
(\%).} Entries are percentages of all sampled rubric--response items.
Agreement is the sum of the two diagonal cells.}
\label{tab:confusion_rar}
\centering
\footnotesize
\setlength{\tabcolsep}{5pt}
\begin{tabular}{@{}lccccc@{}}
\toprule
\multirow{2}{*}{Judge model} &
\multicolumn{2}{c}{Expert: Pass} &
\multicolumn{2}{c}{Expert: Fail} &
\multirow{2}{*}{Agreement}\\
\cmidrule(lr){2-3}\cmidrule(lr){4-5}
& \rar: Pass & \rar: Fail & \rar: Pass & \rar: Fail &\\
\midrule
\modelname{openai}{gpt-oss-120b}  & 23.89 & 14.13 & 13.65 & 48.33 & 72.22\\
\modelname{openai}{GPT-6 Astra}   & 26.82 &  9.42 &  8.96 & 54.80 & 81.62\\
\modelname{claude-ai}{Claude Opus 5} & 27.85 &  8.11 &  8.99 & 55.05 & 82.90\\
\modelname{hunyuan-color}{Hy4 preview}   & 24.99 & 12.86 & 11.75 & 50.40 & 75.39\\
\bottomrule
\end{tabular}
\end{table}

\begin{table}[H]
\caption{\textbf{\method decisions against final expert labels on
PRBench (\%).} Rows give the final expert label and columns the decision of
\method. Entries are percentages of all sampled rubric--response items, the
same denominator as Table~\ref{tab:confusion_rar}. Agreement is the sum of
the three diagonal cells.}
\label{tab:confusion_rar2}
\centering
\footnotesize
\setlength{\tabcolsep}{6pt}
\begin{tabular}{@{}llcccc@{}}
\toprule
\multirow{2}{*}{Judge model} & \multirow{2}{*}{Expert label} &
\multicolumn{3}{c}{\method decision} &
\multirow{2}{*}{Agreement}\\
\cmidrule(lr){3-5}
& & Pass & Fail & Drop &\\
\midrule
\multirow{3}{*}{\modelname{openai}{gpt-oss-120b}}
 & Pass & 26.84 &  1.12 &  0.87 & \multirow{3}{*}{95.91}\\
 & Fail &  0.64 & 55.43 &  0.55 & \\
 & Drop &  0.52 &  0.39 & 13.64 & \\
\midrule
\multirow{3}{*}{\modelname{openai}{GPT-6 Astra}}
 & Pass & 28.93 &  0.97 &  0.12 & \multirow{3}{*}{97.02}\\
 & Fail &  0.28 & 56.77 &  0.36 & \\
 & Drop &  0.09 &  1.16 & 11.32 & \\
\midrule
\multirow{3}{*}{\modelname{claude-ai}{Claude Opus 5}}
 & Pass & 27.63 &  1.86 &  0.27 & \multirow{3}{*}{96.64}\\
 & Fail &  0.33 & 56.04 &  0.54 & \\
 & Drop &  0.05 &  0.31 & 12.97 & \\
\midrule
\multirow{3}{*}{\modelname{hunyuan-color}{Hy4 preview}}
 & Pass & 27.27 &  0.82 &  0.18 & \multirow{3}{*}{98.21}\\
 & Fail &  0.13 & 57.47 &  0.24 & \\
 & Drop &  0.11 &  0.31 & 13.47 & \\
\bottomrule
\end{tabular}
\end{table}

The experts judged $12.57\%$ to $14.55\%$ of items inapplicable, depending on
the judge, and \method matches them on $11.32\%$ to $13.64\%$ of all items.
Most of the gain over \rar, however, comes from fewer Pass/Fail confusions,
which for gpt-oss-120b fall from $27.78\%$ of all items to $1.76\%$. Because
Eq.~\ref{eq:rubric_reliability} measures reproducibility rather than
correctness, these comparisons against human experts provide the evidence
that the retained judgments are also accurate.

\subsection{Human preference evaluation of the trained policies}
\label{app:human_pairwise}

To evaluate the trained policies without rubric scoring, we sampled $100$
held-out questions from each of HealthBench and PRBench and generated one
response per question from the Qwen3-4B policies trained with \rar and with
\method. Experts
compared each pair on overall quality and chose the better response or a tie,
with the response order randomized and the method identities hidden. As
Table~\ref{tab:human_pairwise} shows, experts preferred \method in $72$ of
the $100$ HealthBench pairs and in $79$ of the $100$ PRBench pairs, against
$9$ and $18$ losses.

\begin{table}[!htbp]
\caption{\textbf{Blind pairwise human preference between Qwen3-4B policies
trained with \rar and with \method.} Entries are numbers of pairs from the
perspective of \method. With $100$ pairs per benchmark, the counts equal
percentages.}
\label{tab:human_pairwise}
\centering
\footnotesize
\begin{tabular}{@{}lcccc@{}}
\toprule
Benchmark & Pairs & \method wins & Tie & \rar wins\\
\midrule
HealthBench & 100 & 72 & 19 &  9\\
PRBench     & 100 & 79 &  3 & 18\\
\bottomrule
\end{tabular}
\end{table}

\subsection{Judge-dependent grading stability}
\label{app:judge_stability}

Judgment stability depends on the judge as well as on the rubric and the
response. To measure this dependence, we graded $100$ fixed PRBench policy
outputs four times with each of GPT-6 Astra, Claude Opus 5, and DeepSeek 4.1
Flash \citep{deepseek2026v41flash}. Table~\ref{tab:judge_stability} shows that
stability varies substantially across judges: the mean within-output standard
deviation ranges from $1.45$ to $5.42$ percentage points, and the share of
outputs
scored identically in all four runs ranges from $40\%$ to $76\%$. However, no
judge is fully stable. Even GPT-6 Astra changes the total score of $24\%$ of
outputs across runs.

\begin{table}[H]
\caption{\textbf{Repeated grading of $100$ fixed PRBench outputs, four
runs per judge.} Mean SD is the standard deviation of an output's total
score across the four runs, averaged over outputs. The last column is the
percentage of outputs that received an identical total score in all four
runs.}
\label{tab:judge_stability}
\centering
\footnotesize
\begin{tabular}{@{}lccc@{}}
\toprule
Judge model & Mean score & Mean SD & Identical across 4 runs\\
\midrule
\modelname{openai}{GPT-6 Astra}       & 26.91 & 1.45 & 76\%\\
\modelname{claude-ai}{Claude Opus 5}     & 29.24 & 3.06 & 62\%\\
\modelname{deepseek-color}{DeepSeek 4.1 Flash} & 30.36 & 5.42 & 40\%\\
\bottomrule
\end{tabular}
\end{table}

\section{Detailed Results}
\label{app:detailed_results}

Table~\ref{tab:prbench_detailed} reports PRBench results for both policy
models across domains and difficulty levels.
Tables~\ref{tab:onemillion_4b_detailed} and~\ref{tab:onemillion_30b_detailed}
report \onemillion results across subsets and domains for Qwen3-4B and
\mbox{Qwen3-30B-A3B}, respectively.

\begin{table}[H]
\caption{\textbf{Performance comparison on PRBench by domain and difficulty.}
Each entry reports mean score (\%) $\pm$ standard deviation over three runs.
Boldface marks the best mean for each slice.}
\label{tab:prbench_detailed}
\centering
\begin{tabular}{@{}lcccc@{}}
\toprule
Slice & \rar & \arl & \powthree & \method\\
\midrule
\policyband{5}{Qwen3-4B}\\
Finance Non-Hard & 37.10\sd{0.3120} & 38.19\sd{0.3492} &
                   39.58\sd{0.2720} & \textbf{40.33}\sd{0.2633}\\
Finance Hard     & 24.07\sd{0.4588} & 24.78\sd{0.3379} &
                   25.81\sd{0.2561} & \textbf{26.16}\sd{0.2024}\\
Legal Non-Hard   & 42.52\sd{0.6220} & 43.77\sd{0.5157} &
                   44.65\sd{0.2873} & \textbf{46.22}\sd{0.3887}\\
Legal Hard       & 12.19\sd{0.5414} & 12.55\sd{0.3376} &
                   12.76\sd{0.3965} & \textbf{14.28}\sd{0.1322}\\
\midrule
\policyband{5}{Qwen3-30B-A3B}\\
Finance Non-Hard & 48.15\sd{0.9361} & 49.71\sd{1.0592} &
                   51.20\sd{0.9027} & \textbf{53.38}\sd{0.5051}\\
Finance Hard     & 38.96\sd{1.1789} & 39.02\sd{1.0860} &
                   39.28\sd{0.8673} & \textbf{41.19}\sd{0.6233}\\
Legal Non-Hard   & 46.63\sd{1.1661} & 47.55\sd{0.9423} &
                   49.13\sd{0.9221} & \textbf{50.86}\sd{0.3978}\\
Legal Hard       & 38.39\sd{1.0869} & 39.51\sd{1.0882} &
                   40.23\sd{0.7669} & \textbf{42.24}\sd{0.4932}\\
\bottomrule
\end{tabular}
\end{table}

\begin{table}[H]
\caption{\textbf{Performance comparison of Qwen3-4B on \$OneMillion-Bench
subsets.} Each entry reports mean score (\%) $\pm$ standard deviation over
three runs. Boldface marks the best mean for each slice.}
\label{tab:onemillion_4b_detailed}
\centering
\begin{tabular}{@{}lcccc@{}}
\toprule
Domain & \rar & \arl & \powthree & \method\\
\midrule
\groupband{5}{Global subset}\\
Finance      & 21.10\sd{0.6391} & 22.03\sd{0.7902} &
               22.92\sd{0.5329} & \textbf{25.52}\sd{0.2362}\\
Law          & 22.46\sd{0.9863} & 22.41\sd{0.7126} &
               24.07\sd{0.3298} & \textbf{25.79}\sd{0.5683}\\
Healthcare   & 25.28\sd{0.7565} & 25.22\sd{0.9203} &
               26.38\sd{0.8929} & \textbf{27.82}\sd{0.1661}\\
Natural Sci. & 23.59\sd{0.8546} & 23.63\sd{0.4685} &
               22.94\sd{0.3827} & \textbf{24.28}\sd{0.2816}\\
Industry     & 25.42\sd{0.8723} & 26.10\sd{0.5406} &
               26.79\sd{0.4269} & \textbf{27.14}\sd{0.1899}\\
\midrule
\groupband{5}{CN subset}\\
Finance      & 24.49\sd{0.8674} & 24.72\sd{0.6989} &
               25.35\sd{0.7587} & \textbf{27.79}\sd{0.7855}\\
Law          & 27.28\sd{0.7854} & 28.16\sd{0.8848} &
               29.61\sd{0.7859} & \textbf{30.61}\sd{0.5897}\\
Healthcare   & 24.38\sd{0.8617} & \textbf{24.85}\sd{0.6321} &
               23.85\sd{0.6571} & 24.76\sd{0.5983}\\
Natural Sci. & 21.84\sd{0.5224} & 22.03\sd{0.9051} &
               23.63\sd{0.8762} & \textbf{32.41}\sd{0.7304}\\
Industry     & 27.31\sd{0.9137} & 27.69\sd{0.7587} &
               \textbf{29.77}\sd{0.6159} & 29.25\sd{0.5871}\\
\bottomrule
\end{tabular}
\end{table}

\begin{table}[H]
\caption{\textbf{Performance comparison of Qwen3-30B-A3B on \$OneMillion-Bench
subsets.} Each entry reports mean score (\%) $\pm$ standard deviation over
three runs. Boldface marks the best mean for each slice.}
\label{tab:onemillion_30b_detailed}
\centering
\begin{tabular}{@{}lcccc@{}}
\toprule
Domain & \rar & \arl & \powthree & \method\\
\midrule
\groupband{5}{Global subset}\\
Finance      & 28.36\sd{0.7467} & 28.51\sd{0.4135} &
               29.39\sd{0.5005} & \textbf{31.84}\sd{0.5496}\\
Law          & 29.87\sd{0.8361} & 29.69\sd{0.8315} &
               30.33\sd{0.6224} & \textbf{32.19}\sd{0.6884}\\
Healthcare   & 35.76\sd{0.5328} & 36.53\sd{1.0089} &
               36.72\sd{0.6878} & \textbf{37.26}\sd{0.7581}\\
Natural Sci. & 29.17\sd{0.8433} & 29.83\sd{1.2671} &
               31.52\sd{0.2348} & \textbf{33.97}\sd{0.5731}\\
Industry     & 32.03\sd{0.9443} & 31.97\sd{0.9233} &
               33.45\sd{0.7837} & \textbf{33.89}\sd{0.8206}\\
\midrule
\groupband{5}{CN subset}\\
Finance      & 30.02\sd{0.8522} & 31.24\sd{1.0085} &
               31.97\sd{0.8942} & \textbf{33.43}\sd{0.9182}\\
Law          & 32.79\sd{0.7875} & 32.55\sd{0.8269} &
               33.16\sd{0.8152} & \textbf{34.97}\sd{0.5285}\\
Healthcare   & 36.25\sd{0.6943} & 37.34\sd{0.8571} &
               38.01\sd{0.7732} & \textbf{39.44}\sd{0.3948}\\
Natural Sci. & 29.03\sd{1.0420} & 29.98\sd{0.6882} &
               31.20\sd{0.6942} & \textbf{33.87}\sd{0.4339}\\
Industry     & 33.42\sd{1.0520} & 34.81\sd{0.7846} &
               34.79\sd{0.5283} & \textbf{35.64}\sd{0.6116}\\
\bottomrule
\end{tabular}
\end{table}

\end{document}